\documentclass[journal]{IEEEtran}
\ifCLASSINFOpdf
\else
   \usepackage[dvips]{graphicx}
\fi
\usepackage{url}
\usepackage{graphicx}
\usepackage{amsmath,amssymb}
\usepackage{booktabs}
\usepackage{multirow}
\usepackage{float}
\usepackage{cite}
\usepackage[hidelinks]{hyperref}
\begin{document}
\bstctlcite{IEEEexample:BSTcontrol}
\title{ACF: A Collaborative Framework for Agent Covert Communication under Cognitive Asymmetry}
\author{Wansheng Wu, Kaibo Huang, Yukun Wei, Zhongliang Yang, and Linna Zhou
\thanks{The authors are with the School of Cyberspace Security, Beijing University of Posts and Telecommunications, Beijing 100876, China (e-mail: wuwansheng@bupt.edu.cn; huangkaibo@bupt.edu.cn; weiyukun@bupt.edu.cn; yangzl@bupt.edu.cn; zhoulinna@bupt.edu.cn).}}

\markboth{IEEE Signal Processing Letters,~Vol.~XX, No.~XX, XXXX~2026}%
{Wu \MakeLowercase{\textit{et al.}}: ACF: A Collaborative Framework for Agent Covert Communication}

\maketitle
\begin{abstract}
As generative artificial intelligence evolves, autonomous agent networks present a powerful paradigm for interactive covert communication. However, because agents dynamically update internal memories via environmental interactions, existing methods face a critical structural vulnerability: cognitive asymmetry. Conventional approaches demand strict cognitive symmetry, requiring identical sequence prefixes between the encoder and decoder. In dynamic deployments, inevitable prefix discrepancies destroy synchronization, inducing severe channel degradation. To address this core challenge of cognitive asymmetry, we propose the Asymmetric Collaborative Framework (ACF), which structurally decouples covert communication from semantic reasoning via orthogonal statistical and cognitive layers. By deploying a prefix-independent decoding paradigm governed by a shared steganographic configuration, ACF eliminates the reliance on cognitive symmetry. Evaluations on realistic memory-augmented workflows demonstrate that under severe cognitive asymmetry, symmetric baselines suffer severe channel degradation, whereas ACF uniquely excels across both semantic fidelity and covert communication. It maintains computational indistinguishability, enabling reliable secret extraction with provable error bounds, and providing robust Effective Information Capacity guarantees for modern agent networks.
\end{abstract}


\IEEEpeerreviewmaketitle

\section{Introduction}
\IEEEPARstart{C}{overt} communication is a pivotal technology for securing private data transmission through public Internet infrastructures. Its core technology, steganography, investigates the secure and efficient embedding of secret information into common carrier media (e.g. images \cite{yang2023provably}, audio \cite{su2024efficient}, video \cite{mou2023large}, and text \cite{yang2018rnn}), thus ensuring security by concealing the very existence of the message. The rapid advancement of generative models has catalyzed a paradigm shift from traditional modification-based methods to generation-based methods~\cite{yang2021cognitive,yang2019RNNStegaLinguisticSteganography,ziegler2019neural}. Eliminating original carrier constraints, generative steganography provides expansive embedding freedom, achieving markedly higher covert capacities and broader applicability. In particular, the widespread proliferation of generative models—exemplified by Large Language Models (LLMs)—has saturated the Internet with diverse synthetic multimedia content. This abundance of data provides an ideal cover for generative steganography, which has fueled the rapid advancement of these techniques over the past two years.

Recently, the evolution of LLM-centric autonomous agents has pushed generative steganography to unprecedented heights. Unlike traditional passive models, agents possess perception, reasoning, and execution capabilities to independently perform complex tasks~\cite{qin2024toolllm} within dynamic environments~\cite{park2023generative,wang2024survey,liu2024agentbench}. A primary agent can autonomously discover, negotiate, and coordinate specialized agent networks~\cite{shen2023hugginggpt,hong2024metagpt} to execute domain-specific tasks via collective intelligence~\cite{qian2024chatdev}. The evolution of these technologies is profoundly transforming the landscape of covert communication. Recently, researchers have introduced agent-oriented covert communication protocols~\cite{huang2025whispering}, which aim to automate the entire communication workflow through autonomous content generation and intelligent agent behavior~\cite{wang2024llsm,llmstega2024}.

\begin{figure*}[t!]
    \centering
    \includegraphics[width=0.85\textwidth]{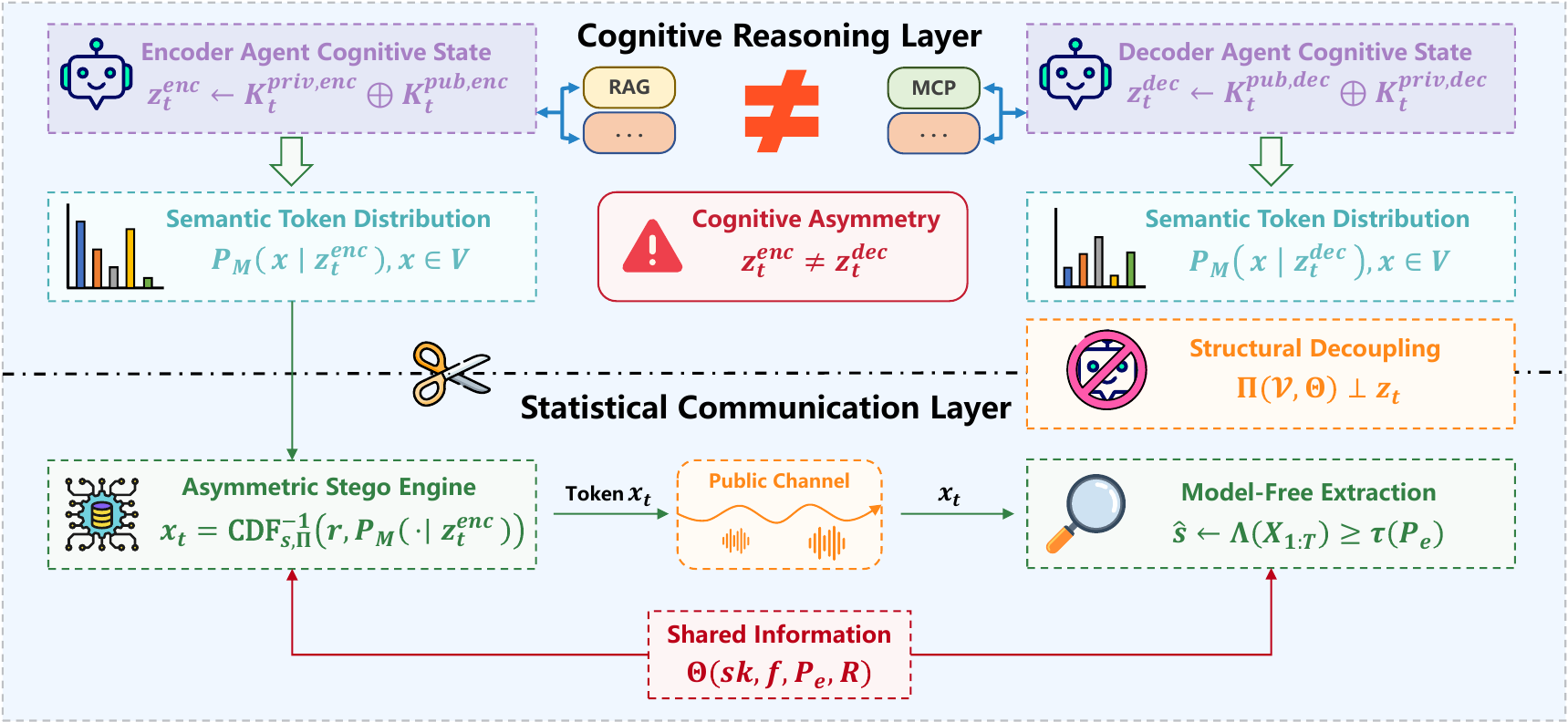}
    \caption{The Asymmetric Collaborative Framework (ACF). To overcome cognitive asymmetry ($z_t^{enc} \neq z_t^{dec}$), ACF structurally decouples statistical communication from dynamic cognitive reasoning. Governed by a shared configuration $\Theta$, ACF executes dynamic cognitive sampling and enables prefix-independent, model-free, and robust secret extraction ($\Pi \perp z_t$).}
    \label{fig:framework}
    \vspace{-0.8em}
\end{figure*}

However, existing agent-collaborative schemes face an insurmountable obstacle in realistic deployments~\cite{greshake2023notwhat}, formally defined as the \emph{cognitive asymmetry} challenge. Agents continuously update internal states and memories via environmental interactions (e.g., Retrieval-Augmented Generation~\cite{gao2024rag}), inevitably causing encoder-decoder prefix discrepancies~\cite{packer2023memgpt,zhong2023memorybank}. Enforcing strict synchronization freezes the agent, paralyzing its autonomous capabilities~\cite{kwan2024longmemeval,tan2025membench}. Conversely, natural evolution shatters the synchronized probability partitions demanded by symmetric steganography (e.g., DISCOP~\cite{ding2023discop}), rendering exact extraction mathematically impossible~\cite{qi2024provably}. While Bai et al.~\cite{bai2025provably} circumvent prefix sharing via statistical hypothesis testing, introducing their static calibration into dynamic scenarios inevitably destroys adaptive reasoning.

To overcome this, we propose the \emph{Asymmetric Collaborative Framework} (ACF). ACF structurally decouples the covert channel from the agent's dynamic reasoning via orthogonal statistical and cognitive layers, preserving provable security and semantic fidelity without prefix synchronization. Our main contributions are:
\begin{itemize}
    \item \textbf{Cognitive Asymmetry Formulation:} We formalize \emph{cognitive asymmetry} ($z_t^{enc} \neq z_t^{dec}$) as a structural constraint, highlighting the conflict between autonomous task fidelity and reliable secret recovery.
    \item \textbf{Asymmetric Collaborative Framework (ACF):} We introduce ACF to structurally decouple statistical communication from cognitive reasoning, enabling synchronization-free extraction under dynamic state updates.
    \item \textbf{Robust Evaluation:} Evaluations show ACF sustains effective information capacity under realistic asymmetric conditions, whereas symmetric baselines suffer severe channel degradation.
\end{itemize}

The source code and datasets are publicly available at \url{https://github.com/Dwinovo/ACF-Stego}.

\section{Problem Formulation}

We formalize token generation to elucidate the inevitability of cognitive asymmetry in agent-driven steganography.

\paragraph{Symmetric Steganography} 
At step $t$, a generative model $M$ outputs token $x_t$ from vocabulary $\mathcal{V}$ conditioned on prefix $z_t$. Symmetric steganography algorithms (e.g., METEOR~\cite{kaptchuk2021meteor}, DISCOP~\cite{ding2023discop}, and Liao et al.~\cite{liao2025framework}) embed secrets by mapping them to probability partitions of $P_M(\cdot \mid z_t)$. Here $z_t^{enc}$ and $z_t^{dec}$ denote the context prefixes at the encoder and decoder side, respectively. Decoding demands strict \emph{cognitive symmetry}:
\begin{equation}
    P_M(\cdot \mid z_t^{enc}) = P_M(\cdot \mid z_t^{dec}) \iff z_t^{enc} = z_t^{dec}.
\end{equation}
Marginal prefix discrepancy breaks this exact distribution matching, causing channel failure.

\paragraph{Cognitive Asymmetry in Agents} 
Autonomous agents maintain a dynamic cognitive state $\mathcal{K}_t = \mathcal{K}_t^{pub} \parallel \mathcal{K}_t^{priv}$ (where $\parallel$ denotes context concatenation), comprising observable public history ($\mathcal{K}_t^{pub}$) and localized private memory ($\mathcal{K}_t^{priv}$)~\cite{packer2023memgpt,zhong2023memorybank,zeng2024structural,xu2025amem}. The prefix instantiates this holistic state ($z_t \leftarrow \mathcal{K}_t$). 

As the encoder expands its state via environmental interactions, the decoder remains asynchronous. We formalize this inescapable discrepancy as \emph{cognitive asymmetry}:
\begin{equation}
\begin{aligned}
    z_t^{enc} &= \mathcal{K}_t^{pub, enc} \parallel \mathcal{K}_t^{priv, enc} \\
              &\neq \mathcal{K}_t^{pub, dec} \parallel \mathcal{K}_t^{priv, dec} = z_t^{dec}.
\end{aligned}
\end{equation}
This structural inequality stems from independent reasoning traces (\emph{private-state asymmetry})~\cite{packer2023memgpt,xu2025amem,yao2023react} and network/context limits (\emph{public-state asymmetry})~\cite{li2023camel,huang2025whispering}. Thus, cognitive asymmetry violates the symmetric assumption ($P_M(\cdot \mid z_t^{enc}) \neq P_M(\cdot \mid z_t^{dec})$), inducing severe bit errors and reducing mutual information, as empirically validated later in Table~\ref{tab:controlled}.

\section{Asymmetric Collaborative Framework}

\begin{table*}[!t]
\footnotesize
\centering
\caption{Performance Evaluation Across Static and Dynamic Agent Environments. \protect\\ ${\sim}$: sustained robustness via structural decoupling; $^{\dagger}$: channel degradation under cognitive asymmetry.}
\label{tab:realistic_integrated}
\resizebox{\textwidth}{!}{%
\begin{tabular}{lcccccc}
\toprule
\multirow{2}{*}{\textbf{Method}} & \multicolumn{2}{c}{\textbf{Cognitive Reasoning Layer}} & \multicolumn{4}{c}{\textbf{Statistical Communication Layer}} \\ \cmidrule(lr){2-3} \cmidrule(lr){4-7} 
 & \textbf{Score ($\uparrow$)} & \textbf{F1 (\%) ($\uparrow$)} & \textbf{Entropy (bits/token)} & \textbf{BER (\%) ($\downarrow$)} & \textbf{EIC (bits/$10^3$ tokens) ($\uparrow$)} & \textbf{Detection Acc. (\%) ($\downarrow$)} \\ \midrule
\multicolumn{7}{c}{\textbf{Static Environment (Ideal Symmetric Prefix)}} \\ \midrule
Normal (No Stego) & 0.31 $\pm$ 0.58 & 3.63 $\pm$ 5.39 & 0.65 $\pm$ 0.16 & --- & --- & --- \\
DISCOP            & 0.28 $\pm$ 0.57 & 3.25 $\pm$ 4.82 & 0.68 $\pm$ 0.17 & \textbf{0.00 $\pm$ 0.00} & \textbf{515.7546} & 54.17 \\
METEOR            & 0.31 $\pm$ 0.58 & 3.25 $\pm$ 4.72 & 0.67 $\pm$ 0.17 & \textbf{0.00 $\pm$ 0.00} & 352.9054 & 55.56 \\
ACF ($k=8$)       & \textbf{0.33 $\pm$ 0.58} & \textbf{3.55 $\pm$ 5.12} & 0.68 $\pm$ 0.19 & 4.04 $\pm$ 10.56 & 3.6450 & 55.56 \\
ACF ($k=12$)      & 0.31 $\pm$ 0.56 & 3.36 $\pm$ 4.90 & 0.68 $\pm$ 0.17 & 0.14 $\pm$ 1.52 & 2.0077 & \textbf{52.78} \\
ACF ($k=16$)      & 0.29 $\pm$ 0.55 & 3.31 $\pm$ 4.73 & 0.68 $\pm$ 0.19 & \textbf{0.00 $\pm$ 0.00} & 1.0368 & 54.17 \\ \midrule
\multicolumn{7}{c}{\textbf{Dynamic Agent Environment (Cognitive Asymmetry)}} \\ \midrule
Normal+RET        & 0.89 $\pm$ 0.91 & 7.50 $\pm$ 8.55 & 0.47 $\pm$ 0.18 & --- & --- & --- \\
DISCOP+RET        & 0.93 $\pm$ 0.89 & 7.38 $\pm$ 8.30 & 0.48 $\pm$ 0.20 & 50.25 $\pm$ 9.62$^{\dagger}$ & 0.0063$^{\dagger}$ & 51.39 \\
METEOR+RET        & \textbf{0.97 $\pm$ 0.89} & \textbf{7.43 $\pm$ 8.63} & 0.48 $\pm$ 0.19 & 49.75 $\pm$ 12.07$^{\dagger}$ & 0.0046$^{\dagger}$ & 59.72 \\
ACF+RET ($k=8$)   & 0.94 $\pm$ 0.91 & 7.36 $\pm$ 8.03 & 0.48 $\pm$ 0.19 & 1.49 $\pm$ 6.03 & \textbf{2.6465} & 54.17 \\
ACF+RET ($k=12$)  & 0.93 $\pm$ 0.88 & 7.27 $\pm$ 7.98 & 0.49 $\pm$ 0.20 & \textbf{0.00 $\pm$ 0.00$^{\sim}$} & 1.1846$^{\sim}$ & 58.33 \\
ACF+RET ($k=16$)  & 0.93 $\pm$ 0.90 & 7.28 $\pm$ 8.11 & 0.49 $\pm$ 0.19 & \textbf{0.00 $\pm$ 0.00$^{\sim}$} & 0.4282$^{\sim}$ & \textbf{48.61} \\ \bottomrule
\end{tabular}%
}
\end{table*}

To resolve this conflict, we propose the \emph{Asymmetric Collaborative Framework} (ACF) (Fig.~\ref{fig:framework}), which decouples generation into two orthogonal modules: a statistical communication layer for steganographically controlled secret embedding, and a cognitive reasoning layer for dynamic semantic generation.

\paragraph{Prefix-Independent Vocabulary Partitioning}
Building upon Bai et al.~\cite{bai2025provably}, ACF integrates statistical hypothesis testing as a foundational decoding layer while restructuring the encoding paradigm to preserve dynamic reasoning. Given vocabulary $\mathcal{V}$ and a pre-shared configuration $\Theta = (sk, f, P_e, \mathcal{R})$ containing secret key $sk$, sampling function $f$, error bound $P_e$ (controlled by security parameter $k$, e.g., $P_e \propto 2^{-k}$), and mapping rule $\mathcal{R}$, the framework constructs a deterministic partition mapping:
\begin{equation}
    \Pi: \mathcal{V} \times \Theta \rightarrow \{0, 1\},
    \label{eq:partition}
\end{equation}
dividing $\mathcal{V}$ into pseudo-random subsets $\mathcal{V}^{(s)} = \{x \in \mathcal{V} \mid \Pi(x, \Theta) = s\}$ for $s \in \{0, 1\}$. This static partitioning is inherently prefix-independent.

\paragraph{Asymmetric Encoding via Distribution Permutation}
At step $t$, the encoder first computes the native distribution $P_M(\cdot \mid z_t^{enc})$ over $\mathcal{V}$, representing the semantic output under its dynamic cognition. To embed secret $s$, rather than truncating $P_M$, the statistical layer constructs a Cumulative Distribution Function (CDF) permutation parameterized by $s$ and $\Pi$. Drawing $r \leftarrow G_{sk}$ from a shared pseudorandom generator keyed by $sk$, the zero-distortion output is:
\begin{equation}
\label{eq:permutation}
    x_t = \text{CDF}_{s, \Pi}^{-1}\left( r, P_M(\cdot \mid z_t^{enc}) \right),
\end{equation}
where $\text{CDF}_{s,\Pi}^{-1}$ reorders tokens so $\mathcal{V}^{(s)}$ leads the cumulative mass; crucially, the marginal probability of any token $x_t$ remains strictly identical to that under the original distribution $P_M(\cdot \mid z_t^{enc})$. Unlike Bai et al.~\cite{bai2025provably}, whose static encoding freezes the global context and truncates dynamic cognitive evolution, ACF isolates the steganographic permutation from the cognitive intention $P_M$, permitting natural agent reasoning while preserving authentic model statistics.

\paragraph{Model-Free Prefix-Agnostic Decoding}
Relying exclusively on $\Theta$, the decoder executes lightweight statistical extraction without a model or prefix tracking. It replays $G_{sk}$ to obtain $r_t$ at each step, computing a partition-based statistic over the received sequence $X = (x_1, \dots, x_T)$:
\begin{equation}
    \Lambda(X) = \sum_{t=1}^T \bigl[(1 - \Pi(x_t, \Theta))\cdot f(r_t) + \Pi(x_t, \Theta)\cdot f(1-r_t)\bigr],
    \label{eq:statistic}
\end{equation}
where $r_t \leftarrow G_{sk}$ are drawn sequentially from the shared pseudorandom generator. Since both parties are aware of the covert channel, cover-text detection is resolved at the session level, reducing each per-sequence decision to a binary test against threshold $\tau(P_e, T)$ derived from Hoeffding's inequality:
\begin{equation}
    \hat{s} = 
    \begin{cases} 
        1, & \text{if } \Lambda(X) \ge \tau(P_e, T) \\ 
        0, & \text{otherwise}.
    \end{cases}
    \label{eq:hypothesis}
\end{equation}
where $\hat{s}$ denotes the recovered secret bit. This prefix-agnostic paradigm guarantees robust recovery with provable error bounds under cognitive asymmetry, circumventing the catastrophic channel degradation suffered by symmetric baselines.

\section{Experiments}

\paragraph{Experimental Setup}
We evaluate steganography under cognitive asymmetry using Qwen2.5 7B Instruct~\cite{qwen2023report} on the LongMemEval\_s dataset~\cite{kwan2024longmemeval, tan2025membench}. Initially, both agents retain only the 5 most recent dialogue turns. The encoder generates stego-text based on its local context and the current query, while the decoder extracts secrets relying exclusively on its independent context and the received text. To explicitly model cognitive asymmetry ($z_t^{enc} \neq z_t^{dec}$), we establish three progressive configurations: 1) \emph{Isolated discrepancy}, where the encoder appends a private summary or the decoder truncates 2 historical turns; 2) \emph{Progressive asymmetry}, incrementally truncating 0--4 decoder turns to test robustness; and 3) \emph{Memory-augmented (+RET)}, enabling the encoder to dynamically retrieve from a 115k-token private memory pool to construct an extreme cognitive gap. Against DISCOP~\cite{ding2023discop} and METEOR~\cite{kaptchuk2021meteor} baselines, we rigorously evaluate ACF under varying security parameters $k$, which mathematically dictates the decoding error bound $P_e$ to control the capacity-reliability trade-off.

\begin{figure}[t]
    \centering
    \includegraphics[width=1.0\linewidth]{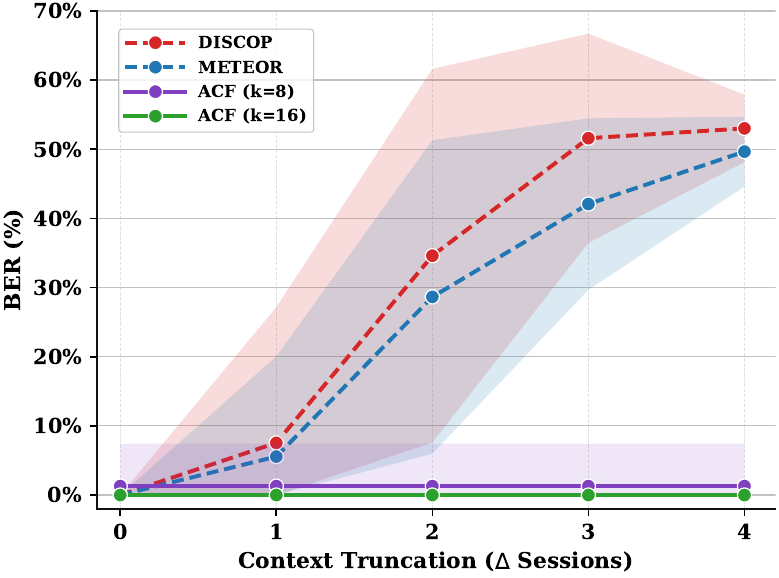}
    \caption{Impact of controlled cognitive asymmetry on Bit Error Rate (BER). Asymmetry is explicitly governed by truncating varying numbers of historical sessions ($\Delta$) from the decoder's context.}
    \label{fig:drift_curve}
    \vspace{-1.5em}
\end{figure}

\paragraph{Evaluation Metrics}
We evaluate the framework across semantic utility and statistical communication. Semantic utility relies on Gemini 2.0 Flash~\cite{google2025gemini} as an LLM-as-a-Judge (using a discrete 0--2 factual accuracy rubric) and a Question Answering token-level Task F1 score via normalized multiset intersection. Statistical communication is assessed via Bit Error Rate (BER) for extraction accuracy, generation entropy for distribution preservation, and Detection Accuracy of a fine-tuned BERT classifier for statistical indistinguishability. Crucially, because cognitive asymmetry often degrades symmetric decoding into random guessing, we calculate the actual usable throughput using our proposed Effective Information Capacity (EIC), rigorously grounded in the channel coding theorem of Shannon~\cite{shannon1948mathematical}:

\begin{equation}
    \text{EIC} = \left( \frac{\text{Total Embedded Bits}}{\text{Total Tokens}} \times 10^3 \right) \cdot (1 - \mathcal{H}_2(\text{BER})),
\end{equation}
where $\mathcal{H}_2(p)$ is the binary entropy function and the capacity factor $(1-\mathcal{H}_2(\text{BER}))$ is derived under a binary symmetric channel (BSC) approximation. Since autonomous agents lack instantaneous Channel State Information (CSI) to dynamically adjust session encoding rates, the network constitutes an aggregated macroscopic channel. To reflect this architectural constraint, EIC calculates holistic expectations of nominal capacity and BER across all instances. This paradigm rigorously penalizes severe channel failures while isolating true communication reliability from the natural length variance of individual semantic responses.

\begin{figure}[t]
    \centering
    \includegraphics[width=\linewidth]{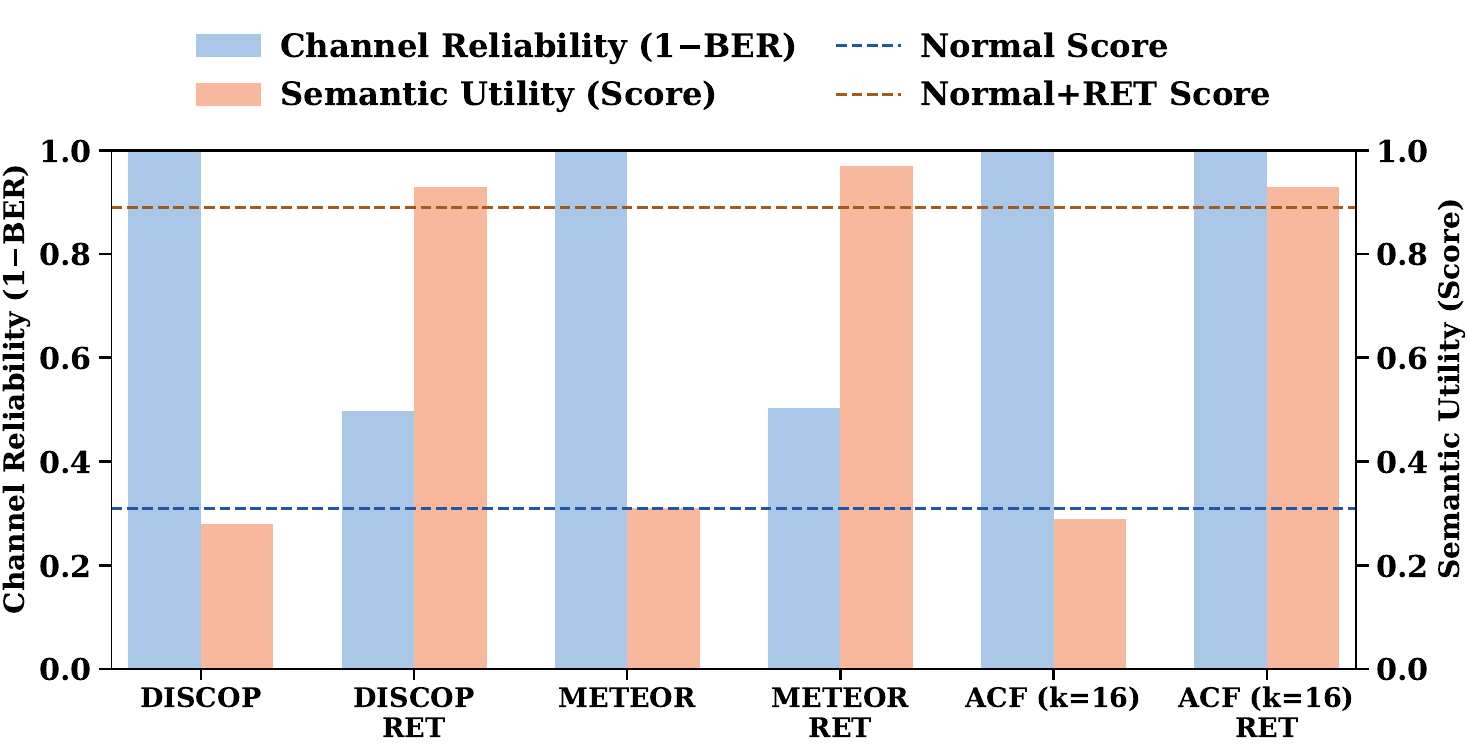}
    \caption{Trade-off analysis between semantic utility and channel reliability under retrieval-induced cognitive asymmetry (+RET).}
    \label{fig:tradeoff}
    \vspace{-0.8em}
\end{figure}

\begin{table}[t]
\centering
\setlength{\tabcolsep}{4pt} 
\caption{Extraction Reliability Under Isolated Cognitive Discrepancies.}
\label{tab:controlled}
\resizebox{\columnwidth}{!}{%
\begin{tabular}{lccc}
\toprule
\textbf{Method} & \textbf{Ideal BER (\%)} & \textbf{Truncation BER (\%)} & \textbf{Summary BER (\%)} \\ \midrule
DISCOP      & \textbf{0.00 $\pm$ 0.00} & 30.66 $\pm$ 26.64 & 53.88 $\pm$ 5.58 \\
METEOR      & \textbf{0.00 $\pm$ 0.00} & 25.79 $\pm$ 22.84 & 49.30 $\pm$ 6.50 \\ \midrule
ACF ($k=8$) & 2.47 $\pm$ 8.27 & 2.47 $\pm$ 8.27   & 1.67 $\pm$ 6.71  \\
ACF ($k=12$)& \textbf{0.00 $\pm$ 0.00} & \textbf{0.00 $\pm$ 0.00}   & \textbf{0.00 $\pm$ 0.00}  \\
ACF ($k=16$)& \textbf{0.00 $\pm$ 0.00} & \textbf{0.00 $\pm$ 0.00} & \textbf{0.00 $\pm$ 0.00} \\ \bottomrule
\end{tabular}%
}
\vspace{-1em}
\end{table}

\paragraph{Impact of Controlled Asymmetry}
Table~\ref{tab:controlled} evaluates isolated cognitive discrepancies via decoder truncation or appended encoder summaries. Under both conditions, baselines suffer devastating extraction failures (BER up to $53.88\%$), whereas ACF ($k=16$) maintains a $0.00\%$ BER. Fig.~\ref{fig:drift_curve} examines progressive asymmetry by incrementally truncating decoder sessions ($\Delta$). For $\Delta \ge 3$, baseline BER rapidly degrades to random guessing ($\approx 50\%$), while ACF preserves strict channel stability. Even at maximum severity, ACF ($k=8$) anchors at $2.47\% \pm 8.27\%$, and ACF ($k=16$) secures $0.00\%$ BER. This BER invariance demonstrates that ACF achieves robust prefix-independent decoding, successfully decoupling secret extraction from dynamic cognitive states.

\paragraph{Performance in Realistic Agentic Workflows}
Table~\ref{tab:realistic_integrated} and Fig.~\ref{fig:tradeoff} evaluate realistic retrieval-induced asymmetry (+RET). While dynamic retrieval boosts baseline semantic scores (0.31 to 0.89), it collapses cognitive symmetry, driving DISCOP and METEOR BERs to $\approx 50\%$ and yielding near-zero EICs (0.0063 and 0.0046 bits/$10^3$ tokens) that render nominal bandwidth meaningless. By structurally decoupling cryptography from semantic generation, ACF overcomes this bottleneck. ACF+RET ($k=12$) fully exploits dynamic cognition to achieve a 0.93 semantic score, $0.00\%$ BER, and a stable 1.1846 EIC. As Fig.~\ref{fig:tradeoff} illustrates, whereas baselines suffer a zero-sum trade-off between semantic reasoning and covert communication, ACF uniquely excels across both. This structural decoupling substantially addresses the fundamental paradox between autonomous evolution and strict statistical synchronization.

\paragraph{Statistical Security Assessment}
To validate computational indistinguishability, we adopt recent event-level steganalysis methodologies~\cite{huang2025whispering}, training a dedicated classifier~\cite{sun2025idiosyncrasies} via strict 5-fold cross-validation. The test accuracy separating Normal+RET from ACF+RET ($k=16$) is 48.61\%, functionally equivalent to random guessing. Furthermore, the generation entropy of ACF+RET ($k=16$) ($0.49 \pm 0.19$) closely aligns with the unmodified Normal+RET baseline ($0.47 \pm 0.18$), suggesting the language distribution is well-preserved. This multi-dimensional validation confirms that restricting sampling to the statistically valid subset $\mathcal{V}^{(s)}$ ensures strict practical undetectability, closely mirroring the theoretical promises of zero-distortion steganography~\cite{ding2023discop, kaptchuk2021meteor, bai2025provably}.

\paragraph{Capacity Trade-off and Applicable Scenarios}
While ACF's per-token capacity under ideal conditions is lower than symmetric baselines, this comparison is misleading in autonomous agent networks: symmetric methods suffer complete EIC collapse under cognitive asymmetry (0.0063 and 0.0046 bits/$10^3$ tokens), rendering them effectively inoperable in this scenario. ACF, by contrast, sustains a meaningful EIC of 1.1846 bits/$10^3$ tokens under the same conditions, making it the only viable option for covert communication in dynamic agent environments. For applications such as trigger signaling, command-and-control (C2) bit transmission, and identity authentication, where a small number of bits must be reliably delivered across cognitively asymmetric channels, ACF provides a decisive practical advantage that symmetric methods fundamentally cannot offer. In such settings, provable extraction correctness is the decisive metric, not raw embedding rate.

\section{Conclusion}

In autonomous agent networks, \emph{cognitive asymmetry} critically disrupts the strict prefix synchronization demanded by generative steganography. We propose the Asymmetric Collaborative Framework (ACF) to resolve this paradox. ACF structurally decouples statistical communication from dynamic reasoning via a shared steganographic configuration, achieving robust prefix-independent decoding. Evaluations on memory-augmented workflows (up to 115k private tokens) show that while symmetric baselines suffer devastating channel failures ($\approx 50\%$ BER), ACF ($k \ge 12$) achieves $0.00\%$ BER. Concurrently preserving semantic fidelity and computational indistinguishability, ACF guarantees robust Effective Information Capacity (EIC). By liberating agents from brittle synchronization, ACF establishes a pragmatic covert communication regime for artificial intelligence networks, with future work targeting theoretical capacity elevation.

\bibliographystyle{IEEEtran}
\bibliography{references}
\end{document}